\pdfoutput=1
\documentclass[
twocolumn,
]{ceurart}

\begin{document}

\copyrightyear{2021}
\copyrightclause{Copyright for this paper by its authors.
  Use permitted under Creative Commons License Attribution 4.0
  International (CC BY 4.0).}

\conference{CDCEO 2021: 1st workshop on Complex Data Challenges in Earth Observation, November 1, 2021, Virtual}

\title{A Variational U-Net for Weather Forecasting}

\author[1]{Pak Hay Kwok}[%
email={pak_hay_kwok@hotmail.com},
url=https://github.com/ivans-github/,
]

\author[2]{Qi Qi}[%
email=qiq208@gmail.com,
url=https://github.com/qiq208/,
]

\address[1]{pak\_hay\_kwok@hotmail.com}
\address[2]{qiq208@gmail.com}

\begin{abstract}
  Not only can discovering patterns and insights from atmospheric data enable more accurate weather predictions, but it may also provide valuable information to help tackle climate change. Weather4cast is an open competition that aims to evaluate machine learning algorithms’ capability to predict future atmospheric states. Here, we describe our third-place solution to Weather4cast. We present a novel Variational U-Net that combines a Variational Autoencoder’s ability to consider the probabilistic nature of data with a U-Net’s ability to recover fine-grained details. This solution is an evolution from our fourth-place solution to Traffic4cast 2020 with many commonalities, suggesting its applicability to vastly different domains, such as weather and traffic.
  
  The code for this solution is available at \url{https://github.com/qiq208/weather4cast2021_Stage1}
\end{abstract}


\maketitle

\section{Introduction}

Meteorological satellites around the globe are constantly gathering a trove of data about the atmosphere. However, the high-dimensionality nature of atmospheric data makes it challenging to analyse, hindering the discovery of valuable insights. With the advent of machine learning methods, it is believed these methods can help better understand atmospheric data. To evaluate the applicability of such techniques to atmospheric data, Weather4cast \cite{W4cWeb} by the Institute of Advanced Research in Artificial Intelligence is an open competition that challenges its participants to develop algorithms to predict the future states of the atmosphere over specific regions.

\begin{figure*}
  \centering
  \includegraphics[width=\textwidth]{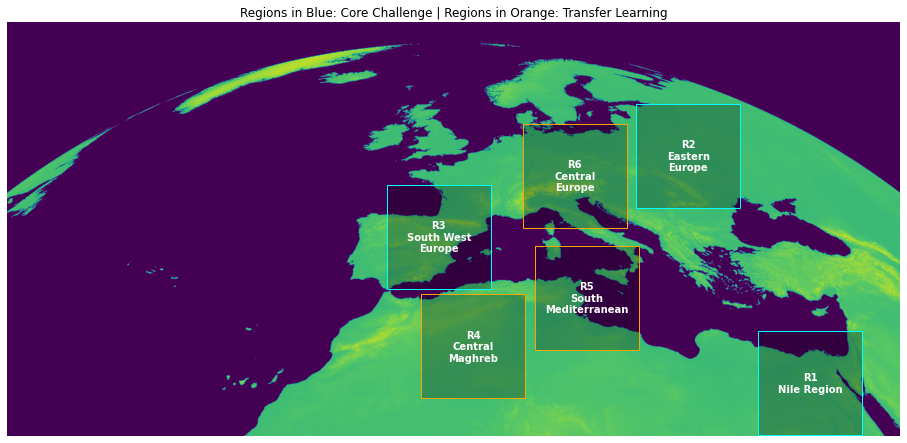}
  \caption{Weather4cast regions}
  \label{fig:w4c_regions}
\end{figure*}

The Weather4cast dataset \cite{W4cGithub} is obtained from Meteosat geostationary meteorological satellites operated by EUMETSAT for the period from February 2019 to February 2021. The Meteosat images are processed by NWC SAF software into weather products. The weather products of interest are: Cloud Top Temperature and Height (CTTH), Convective Rainfall Rate (CRR), Automatic Satellite Image Interpretation - Tropopause Folding detection (ASII-TF), Cloud Mask (CMA), and Cloud Type (CT). Each of these weather products is recorded in 15-minute intervals and consists of multiple channels. Each channel is in the format of an image of shape 256x256 pixels, with each pixel covering an area of about 4x4 km. The regions of interest are illustrated in Figure \ref{fig:w4c_regions}; regions R1-3 correspond to the core challenge in which training, validation and test data are provided, while regions R4-6 correspond to the transfer learning challenge in which only the test data are provided. In addition, static information, such as altitude, latitude and longitude, are also given for all regions.

Weather4cast demands an algorithm that can return the atmospheric states over the defined regions for the next 8 hours (32-off 15-minute intervals) given an hour (4-off 15-minute intervals) worth of data. While only 4 target variables are required, namely $temperature$ (a channel of CTTH), $crr\_intensity$ (a channel of CRR), $asii\_turb\_trop\_prob$ (a channel of ASII-TF) and $cma$ (a channel of CMA), any channels of the weather products or static information of the regions can be used as input variables.

This work describes a novel Variational U-Net solution which achieved third place in both the core and transfer learning challenges of Weather4cast. This Variational U-Net can be viewed as a U-Net with a Variational Autoencoder (VAE) style bottleneck, or as a VAE with U-Net style skip connections. The intuition behind this architecture is to combined VAE’s ability to consider the probabilistic nature of data with U-Net’s ability to recover fine-grained details.

\section{Related work}

Weather4cast can be viewed as a video frame prediction problem, in which the inputs are the first 4 frames of a video, and the outputs are the subsequent 32 frames. This format of the problem is identical to that of Traffic4cast \cite{kreil2020surprising, kopp2021traffic4cast}. Overlooking the difference in domains between Weather4cast and Traffic4cast, the two competitions can be considered the same, hence solutions for Traffic4cast should be somewhat transferable to Weather4cast. A range of algorithms, including U-Net, LSTM and Graph Neural Network were proposed for Traffic4cast \cite{qi2020traffic4cast, martin2019traffic4cast}, yet various flavours of U-Net dominated the competition in both 2019 and 2020, with all winning teams adopting U-Nets in their final solutions \cite{qi2020traffic4cast, choi2020utilizing}. Thus, it is sensible to consider U-Net-based solutions for Weather4cast.

While the formats of Weather4cast and Traffic4cast are equivalent, the differences in the underlying domains cannot be ignored. Specifically, weather is considered more random than traffic. Multiple scenarios are possible given a set of observations, and this inherent randomness needs particular attention, as it is not compatible with the deterministic nature of a typical U-Net. Segmentation of medical images also suffers from intrinsic ambiguities. To handle these ambiguities, Kohl et al. \cite{kohl2018probabilistic} proposed a Probabilistic U-Net, a combination of a U-Net with a conditional VAE, capable of producing an unlimited number of hypotheses from a set of inputs. Myronenko \cite{myronenko20183d} also proposed a different way to combine a U-Net with a VAE, which a VAE was applied to regularise a shared encoder. His solution was proven successful and won first place in the Multimodal Brain Tumour Segmentation Challenge (BraTS) in 2018.

\section{Methods}
\subsection{Model architecture}

Given the similarities between Weather4cast and Traffic4cast, the main structure of the proposed Variational U-Net largely resembles the authors’ fourth-place solution to last year’s Traffic4cast \cite{qi2020traffic4cast}. The encoder is made up of Dense Blocks connected by 2D Max Pooling. Each Dense Block consists of 4 repeats of 2D Convolution, ELU \cite{clevert2015fast}, Group Normalisation \cite{wu2018group} and 2D Dropout \cite{tompson2015efficient}, followed by another 2D Convolution and ELU. Different to the encoder, the decoder consists of repeats of 2D Transposed Convolution, ELU, 2D Convolution, ELU, Group Normalisation and 2D Dropout. The encoder and the decoder are joined by skip connections.

Inspired by the works of Kohl et al. \cite{kohl2018probabilistic} and Myronenko \cite{myronenko20183d}, the bottleneck of the Variational U-Net, the part which connects the end of the encoder to the start of the decoder, is replaced with one that is typically found in VAE. At the end of the encoder, the input is reduced to 2 vectors of size 512, representing the means and standard deviations of the latent variables. With the assumption that the latent variables are Gaussian, a sample is drawn, and the drawn vector is reconstructed into an image which is then passed through the decoder.

\begin{figure*}
  \centering
  \includegraphics[width=\textwidth]{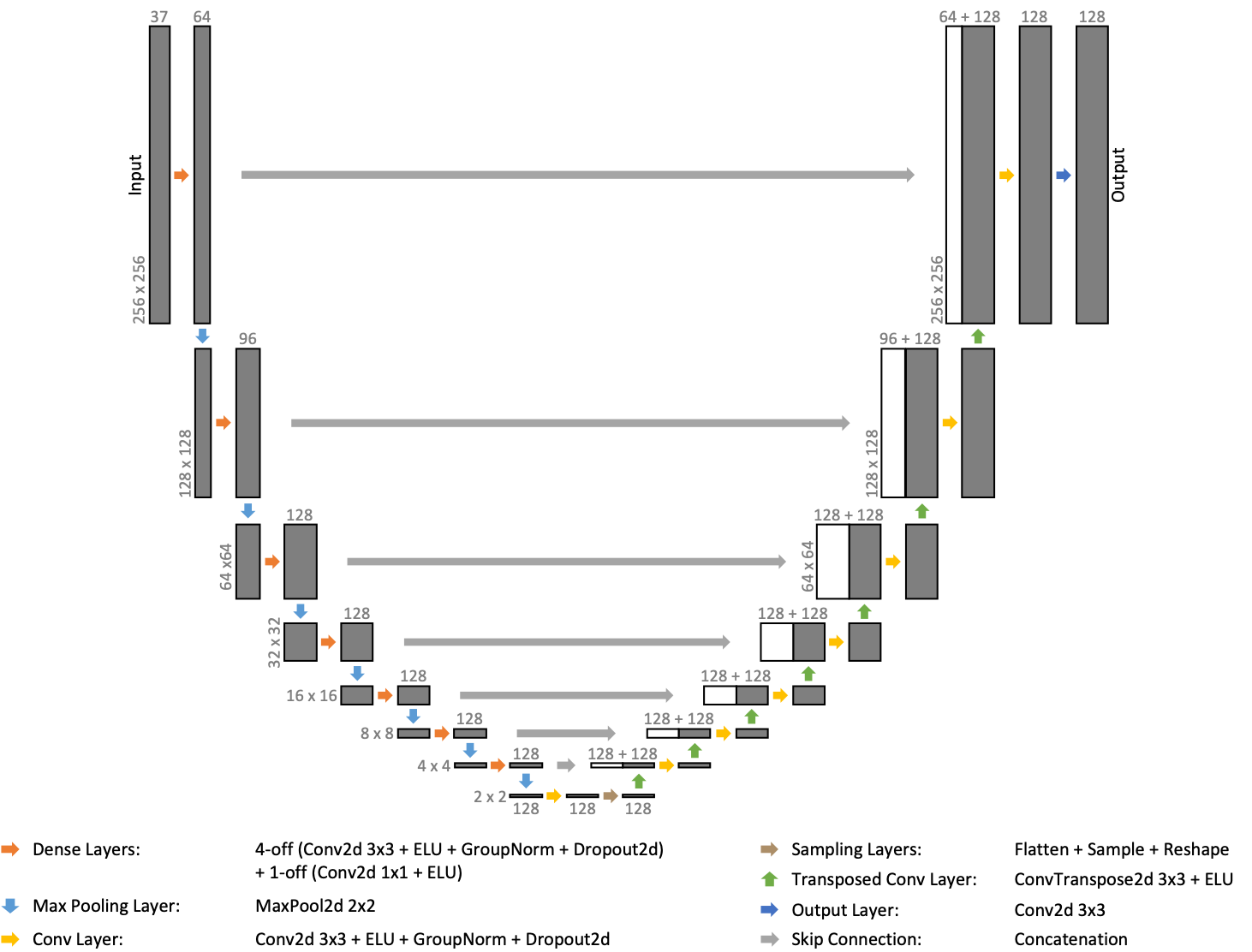}
  \caption{Variational U-Net architecture}
  \label{fig:vunet}
\end{figure*}

The architecture of the Variational U-Net is shown in Figure \ref{fig:vunet}. 

\subsection{Inputs and target variables}

Similar to the authors’ Traffic4cast solution \cite{qi2020traffic4cast}, the temporal dimension of the input tensor is combined with the channel dimension, resulting in the number of input channels of 4*8. Furthermore, since it seems intuitive that weather patterns are dependent on geographical location, the static features of altitude, latitude and longitude are appended, resulting in an additional 3 input channels. As such, the final number of input channels to the Variational U-Net is 4*8+3=35. Finally, the model is designed to predict all 32 output frames in one go, resulting in the number of output channels being 32*4=128. Furthermore, any missing data has been zero-filled.

A series of experiments were performed to find the most effective set of input features, and the validation set was used to evaluate the performance of each feature set. The resulting input feature set is listed in Table \ref{tab:feats_used}, and those rejected are summarised in Table \ref{tab:feats_unused}. 

\begin{table*}
  \caption{Summary of input features and target variables}
  \label{tab:feats_used}
  \begin{tabular}{CCC}
    \toprule
    Feature & Target Variable & Description\\
    \midrule
    $temperature$ & Yes & Combined cloud top and ground temperature\\
    $ctth\_pres$ & No & Cloud top pressure\\
    $crr\_intensity$ & Yes & Convective rainfall rate intensity in mm/h\\
    $crr\_accum$ & No & Convective rainfall rate hourly accumulations\\
    $asii\_turb\_trop\_prob$ & Yes & Probability of occurrence of tropopause folding\\
    $cma$ & Yes & Cloud mask\\
    $ct$ & No & Cloud type\\
    $ctth\_tempe$ mask & No & A mask showing pixel locations containing cloud top temperature measurements\\
  \bottomrule
\end{tabular}
\end{table*}

\begin{table*}
  \caption{Summary of input features not used in the final model}
  \label{tab:feats_unused}
  \begin{tabular}{cc}
    \toprule
    Feature & Description\\
    \midrule
    $ctth\_alt$ & Cloud top altitude\\
    Linear interpolation of $temperature$ & Using linear interpolation to fill in missing $temperature$\\
    Linear interpolation of $ctth\_pres$ & Using linear interpolation to fill in missing $ctth\_pres$\\
  \bottomrule
\end{tabular}
\end{table*}

\subsection{Loss function}

The loss function consists of 2 terms:
\begin{equation}
  \label{eq:loss}
  L = L_{L2} + 80 * L_{KL}
\end{equation}

$L_{L2}$ is a modified mean squared error, it takes into account missing values and the difference in scale of the 4 target variables:
\begin{equation}
  L_{L2} = \frac{1}{32 \times 4} \sum_{t=1}^{32} \sum_{v \in V} \frac{w_v}{P_{t,v}} \sum_{p=1}^{P_{t,v}} (y_{t,v,p} - \hat{y}_{t,v,p})^{2}
\end{equation}
where $V = \{temperature, crr\_intensity, cma,\\ asii\_turb\_trop\_prob\}$, $P_{t,v}$ is the total number of non-missing pixels for a given target variable $v$ at a given time $t$ and $w_v$ is the target variable weighting:
\begin{displaymath}
  w_v = 
  \begin{cases}
    31.610, & v = temperature\\
    4139.4, & v = crr\_intensity\\
    5.2191, & v = cma\\
    142.17, & v = asii\_turb\_trop\_prob
  \end{cases}
\end{displaymath}

$L_{KL}$ is the KL divergence between the estimated Gaussian distribution $N(\mu, \sigma^2)$ and a prior distribution $N(0, 1)$:
\begin{equation}
  L_{KL} = \frac{1}{2} \sum_{i=1}^{512} \mu_i^2 + \sigma_i^2 - \log \sigma_i - 1
\end{equation}

The $L_{KL}$ factor of 80 in Equation \ref{eq:loss} was determined empirically to balance the relative importance of the two terms in the loss function.

\subsection{Optimisation}

The Variational U-Net is trained using the Adam optimiser with Cyclic Cosine Annealing described by Loshchilov et al. \cite{loshchilov2016sgdr}. The training process is split into cycles, with each cycle consisting of 2 epochs. At each cycle, the learning rate is first set to a maximum of 2e-4, then is reduced following a cosine annealing schedule. Resetting the learning rate at the beginning of each cycle perturbs the models and encourages them to explore different basins of attraction. The training is continued until an additional cycle failed to return a better validation score.

Using a batch size of 12, the final model was first trained for 6 cycles (12 epochs) on the training data, then it was further trained for an additional cycle (2 epochs) on both the training and validation data.

\subsection{Regularisation}

From initial experiments, it became apparent that controlling overfitting of the model to the training data was a key to success in both the core and transfer learning challenges. Hence, several regularisation strategies were employed. Within the model itself, the move to the Variational U-Net from a traditional U-Net, combined with the introduction of dropout layers throughout the encoder and decoder, both aimed to improve the generalisation of the model. To expose the model to as much variation in input as possible, a single model was used for all regions in the competition and trained on all available training data. Furthermore, for the final leaderboard submission, the model was further trained for another cycle on all the validation data available.

\section{Results}

The majority of experimentation on the design of features and model architecture was conducted on single regions to allow for quicker feedback and learning. However, the final model was trained on data from all regions, so there is a risk that some of the decisions made might not be optimum for a model trained on data from all regions. Results from the main experiments can be found in Appendix \ref{tab:experiments}.

Final experiments on all three regions were conducted, and models were evaluated based on either the test learderboard or the final leaderboard. It is worth noting that the test leaderboard allowed multiple submissions and was open up to the final week of the competition. In the final week, the final leaderboard was opened and competitors were only allowed three submissions. The results of the submissions can be found in Table \ref{tab:results}.

\begin{figure*}
  \centering
  \includegraphics[scale=0.75]{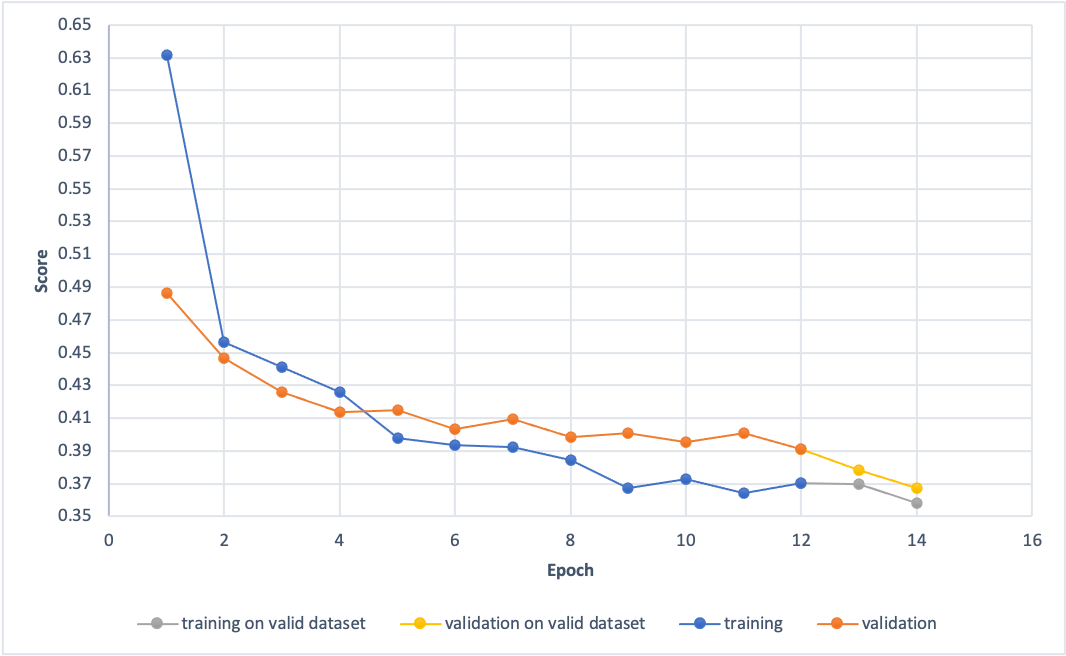}
  \caption{Training history of the final model}
  \label{fig:loss}
\end{figure*}

The competition is based on the final leaderboard scores and the final model resulted in a third-place finish for both the core and transfer learning challenges. The training history of the final model is shown in Figure \ref{fig:loss}, highlighting the loss progression during both the normal training phase, as well as the additional cycle training on the validation data.

\begin{table*}
  \caption{Summary of leaderboard scores for final models}
  \label{tab:results}
  \begin{tabular}{p{4.5cm}cccccc}
    \toprule
    \multirow{3}{*}{Model} & \multicolumn{3}{c}{Core Challenge} & \multicolumn{2}{c}{Transfer Learning Challenge} \\
    \cline{2-6}
     & \multirow{2}{*}{Validation} & Test  & Final  & Test  & Final \\
     &  &  Leaderboard &  Leaderboard &  Leaderboard &  Leaderboard\\
    \midrule
    Mean baseline & - & 0.8822 & - & - & -\\
    IARAI U-Net baseline \cite{W4cGithub} & - & 0.6689 & - & 0.6111 & -\\
    \\
    One model per region & - & 0.5095 & - & - & -\\
    Single model & 0.3912 & 0.4977 & 0.5140 & 0.4878 & 0.4711\\
    Single model + linear interpolation of $temperature$ & 0.3887 & - & 0.5218 & - & -\\
    Single model + training on validation data & - & - & 0.5102 & - & 0.4670\\
  \bottomrule
\end{tabular}
\end{table*}

\section{Discussion}

Although various U-Net architectures were explored, it was interesting to observe that the final architecture was very similar to the architecture used for Traffic4cast \cite{qi2020traffic4cast}. The only changes were moving to max pooling from average pooling, the addition of dropout layers and the adoption of the VAE style bottleneck. The authors would be interested in exploring whether these improvements would also read back across to the traffic prediction task.

In terms of feature engineering, the experiments showed that the inclusion of some extra features (e.g. cloud top pressure) improved predictive capability, whereas others (e.g. cloud top altitude) did not. It was found that linearly interpolating temperature provided an improvement to the validation score, however, this did not read across to the final leaderboard score. The authors still believe that strategies to compute missing data is an interesting area for further work. 

Perhaps most surprisingly was the benefit gained from training a single model on data from all regions instead of individual models for each region. The model trained on all regions displayed a significant improvement in the test leaderboard score (\textasciitilde2.3\%) over individually trained models. This finding suggests that that the model may continue to improve its general predictive ability for any region with the addition of more training data. This hypothesis was further supported as training on the validation data further improved the final leaderboard score for both core and transfer learning challenges.

\section{Conclusion}

Weather4cast provided the opportunity to explore the use of machine learning techniques to the age-old problem of weather forecasting. Furthermore, the similarity of format to Traffic4cast also provided the chance to investigate how transferable machine learning models can be across vastly different domains. After experimenting with various U-Net architectures, the final model was very similar to the authors’ Traffic4cast model. The main differences being changes to suppress overfitting, i.e. moving to the Variational U-Net model and inclusion of dropout layers throughout. The authors also found that training on data from all regions in one model outperformed training individual models on each region for both the core and transfer learning challenges. This suggests that the model prediction for all regions can be improved by training on more data. 

\bibliography{references}

\appendix

\section{Experiments on R1}
\label{tab:experiments}
Table \ref{tab:exp_results} details some of the experiments done on R1 to explore which input features should be included in the final model. All these experiments were done using the training and validation data provided. The underlying assumption was that the results from these experiments would read across to the final leaderboard.
\setcounter{table}{0}
\renewcommand{\thetable}{A\arabic{table}}
\begin{table}[h]
  \tiny
  \caption{Summary of experimental results on R1}
  \label{tab:exp_results}
  \begin{tabular}{lccccc}
    \toprule
    Experiment & Base & 1 & 2 & 3 & 4\\
    \midrule
    $ctth\_pres$ & - & - & Yes & Yes & Yes \\
    $crr\_accum$ & - & Yes & Yes & Yes & Yes \\
    $ct$ & - & - & Yes & Yes & Yes \\
    $ctth\_tempe$ mask & - & - & - & Yes & Yes \\
    $ctth\_alt$ & - & - & Yes & - & - \\
    Interpolated $ctth\_tempe$ & - & - & - & - & Yes \\
    \\
    Epoch & 20 & 27 & 32 & 24 & 20 \\
    Training score & 0.2247 & 0.2155 & 0.2091 & 0.2087 & 0.2229 \\
    Validation score & 0.1933 & 0.1935 & 0.1894 & 0.1879 & 0.1889 \\
    \bottomrule
  \end{tabular}
\end{table}

\end{document}